\pdfoutput=1

\documentclass[11pt]{article}

\usepackage[preprint]{acl}

\usepackage{times}
\usepackage{latexsym}
\usepackage{amsmath}
\usepackage{amsfonts}
\usepackage{longtable}
\usepackage{booktabs}  
\usepackage{array}     
\usepackage{graphicx}  
\usepackage[T1]{fontenc}

\usepackage{microtype}

\usepackage{inconsolata}

\usepackage{graphicx}
\usepackage{stmaryrd}
\usepackage{inconsolata}
\usepackage{microtype}
\usepackage{enumitem}
\usepackage{inconsolata}
\usepackage{todonotes}
\usepackage[normalem]{ulem}
\usepackage{booktabs}
\usepackage{subcaption}
\usepackage{hyperref}
\usepackage{url}
\usepackage{graphicx}
\usepackage{multirow}
\usepackage{arabtex}
\usepackage{utf8}
\setcode{utf8}
\usepackage{kotex}
\usepackage{comment}
\usepackage[linesnumbered, ruled, vlined]{algorithm2e}

\usepackage{listings}
\usepackage{color}

\usepackage{pifont}
\definecolor{blue}{rgb}{0,0, 0.6}
\definecolor{dkgreen}{rgb}{0,0.6,0}
\definecolor{gray}{rgb}{0.5,0.5,0.5}
\definecolor{mauve}{rgb}{0.58,0,0.82}
\definecolor{mauve}{rgb}{0,0,0}
\definecolor{black}{rgb}{0,0,0}

\usepackage[normalem]{ulem}

\usepackage{arabtex}
\usepackage{utf8}
\setcode{utf8}

\usepackage{soul}
\definecolor{tri}{rgb}{.25,.88,.82}
\definecolor{lilac}{rgb}{0.85,0.64,0.85}
\definecolor{atomictangerine}{rgb}{1.0, 0.6, 0.4}
\definecolor{lightblue}{rgb}{0.53, 0.81, 0.98} 

\newcommand{\llens}{\emph{LlamaLens}}

\usepackage{algorithmicx}
\usepackage{algpseudocode}

\makeatletter
\algrenewcommand\ALG@beginalgorithmic{\footnotesize} 
\makeatother
\title{LlamaLens: Specialized Multilingual LLM for \\Analyzing News and Social Media Content}

\author{Mohamed Bayan Kmainasi$^1$\thanks{The contribution was made while the author was interning at the Qatar Computing Research Institute.}\textsuperscript{$\dagger$}, Ali Ezzat Shahroor$^2$\thanks{~~Equal contribution.}, Maram Hasanain$^2$, \\{\bf Sahinur Rahman Laskar$^3$, Naeemul Hassan$^4$, Firoj Alam$^2$}\\
$^1$Qatar University, Qatar, 
$^2$Qatar Computing Research Institute, Qatar\\
$^3$UPES, India,
$^4$University of Maryland, USA\\
Mk2314890@qu.edu.qa, \{ashahrrour, sahinurlaskar.nits\}@gmail.com,\\nhassan@umd.edu,
\{mhasanain, fialam\}@hbku.edu.qa
}

\begin{document}
\maketitle
\begin{abstract}
Large Language Models (LLMs) have demonstrated remarkable success as general-purpose task solvers across various fields. However, their capabilities remain limited when addressing domain-specific problems, particularly in downstream NLP tasks. Research has shown that models fine-tuned on instruction-based downstream NLP datasets outperform those that are not fine-tuned. While most efforts in this area have primarily focused on resource-rich languages like English and broad domains, little attention has been given to multilingual settings and specific domains. To address this gap, this study focuses on developing a specialized LLM, \textit{LlamaLens}, for analyzing news and social media content in a multilingual context. To the best of our knowledge, this is the \textit{first attempt} to tackle both domain specificity and multilinguality, with a particular focus on news and social media. Our experimental setup includes 18 tasks, represented by 52 datasets covering Arabic, English, and Hindi. We demonstrate that \textit{LlamaLens} outperforms the current state-of-the-art (SOTA) on 23 testing sets, and achieves comparable performance on 8 sets. We make the models and resources publicly available for the research community.\footnote{\url{https://huggingface.co/QCRI}}
\end{list}
\end{abstract}

\section{Introduction}
\label{sec:intro}


\begin{figure}[t]
    \centering    
    \includegraphics[scale=0.3]{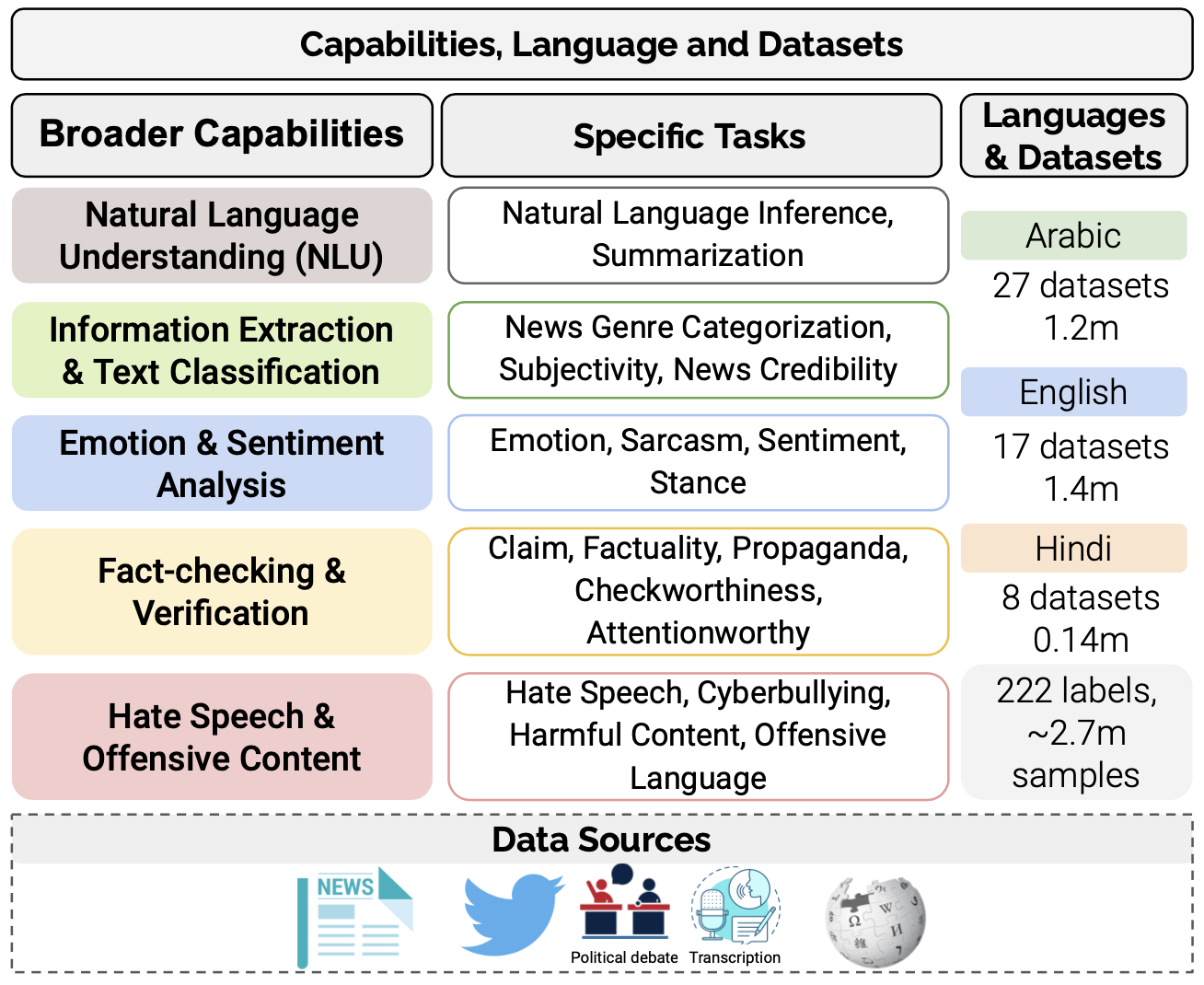}
    \caption{Capabilities, tasks and associated datasets in \llens{}. 
    }
    \label{fig:llamalens_capablities_tasks_datasets}
    \vspace{-0.4cm}
\end{figure}

LLMs have significantly advanced the field of AI, demonstrating capabilities in solving downstream NLP tasks and exhibiting knowledge understanding and cognitive abilities~\cite{touvron2023llama,mousi2024aradice}. However, extending these capabilities with more domain-specific knowledge and achieving higher accuracy requires domain specialization. This entails customizing general-purpose LLMs with domain-specific data, augmented by relevant domain knowledge~\cite{ling2023domain}. 

One prominent area where LLMs can be customized with specialized knowledge is the news and social media analysis. Since their emergence, the use of AI in news production, analysis, video scripting, copyediting, translation, and dissemination has grown significantly~\cite{journalmedia5020039,simon2024artificial}. In addition to news production and dissemination, a closely related area is social media content analysis~\cite{zeng2024large,liu2024using,franco2023analyzing,hasanain2024large}. This growing range of applications creates a strong demand for specialized LLMs to support journalists, fact-checkers, communication specialists and social media users. 

There has been an attempt to develop a tool\footnote{\url{https://newsgpt.ai/}} based on ChatGPT to support journalists in news production and delivery~\cite{hireche2023newsgpt}; however, it relies solely on ChatGPT for language understanding and response generation in response to reporters' queries. Other efforts to support journalists include tools for creating news reels~\cite{wang2024reelframer}, classifying news frames using GPT~\cite{alonso2023framing}, generating image captions for news articles~\cite{anagnostopoulou2024enhancing}, news recommendation systems~\cite{wang2024cherryrec}, and specialized models for the news media and business sectors~\cite{bao2024harnessing}. However, little to no attention has been given to developing specialized models for news and social media content analysis (see related work in Section \ref{ssec:specialized_llm}). Such models are crucial for tasks like identifying whether a news article or social media post contains a claim, assessing its factual accuracy, determining its relevance for fact-checking, and evaluating whether the content is offensive, incites hate, or requires moderation.  

To address this gap, we focused on developing specialized LLMs by fine-tuning an existing model to better support journalists, fact-checkers, communication specialists, and social media analysts. As illustrated in Figure \ref{fig:llamalens_capablities_tasks_datasets}, our goal was to equip LLMs with a range of specialized capabilities across multiple languages. 
Our main contributions are summarized below. 
\begin{itemize}[noitemsep,topsep=0pt,leftmargin=*,labelsep=.5em]
\item We develop and release a specialized LLM, \llens{}, which covers 5 broader capabilities, associated with 18  tasks and 52 datasets in three languages: Arabic, English, and Hindi.
\item We develop and release an instruction-following dataset, 
developed using a semi-supervised approach.
\item We explore various data shuffling techniques, based on language, dataset, and task, during training and present our findings. 
\item We present detailed experimental results comparing with \textit{(i)} Llama-3.1-8B-Instruct model, \textit{(ii)} quantized models trained over data with different shuffling,
and \textit{(iii)} state-of-the-art baselines using dataset-specific metrics.
\end{itemize}

Our findings suggest that \textit{(i)} \llens{} acquires domain- and language-specific knowledge, highlighting the importance of specialized models, \textit{(ii)} considerably smaller versions of the model (resulting from fine-tuning with quantization) acquire that knowledge in comparison to the un-finetuned model, showing significantly better performance, and \textit{(iii)} comparison to the SOTA suggests that there is still a room for improvement.

\section{Related Work}


\subsection{LLMs for Journalism}

Recent studies have explored the intersection of LLMs, journalism, and social media, shedding light on both the opportunities and challenges of integrating AI into news reporting~\cite{cheng2024journalism,10.1145/3544548.3580907,hasan-etal-2024-llm,DBLP:journals/aim/QuinonezM24,nishal2024envisioningapplicationsimplicationsgenerative,ding2023harnessingpowerllmsevaluating}.  For example,~\citet{brigham2024breakingnewscasestudies} and ~\citet{breazu2024chatgpt} examined the use of LLMs like GPT-4 in journalistic workflows, focusing on the ethical, and quality implications and generating narratives. LLMs have also been integrated in news production, focusing on its benefits and ethical challenges~\cite{journalmedia5020039}. One key challenge of using generic LLMs in journalism is their tendency to generate false or misleading information~\cite{cheng2024journalism,augenstein2023factualitychallengeseralarge}, a phenomenon known as hallucination. 
\citet{nishal2024envisioningapplicationsimplicationsgenerative} also highlighted key concerns, including hallucinations, factual inaccuracies, and the potential threat to journalistic objectivity. To address these issues, they proposed a value-sensitive design approach, advocating for AI systems that offer transparent explanations, explicitly represent uncertainty, and give journalists more control over the generated content.

Bloomberg has integrated LLMs into its news production processes~\cite{DBLP:journals/aim/QuinonezM24}, aiming to enhance automation while preserving essential journalistic principles such as accuracy and transparency. Similarly, ~\citet{ding2023harnessingpowerllmsevaluating} examined the role of LLMs in human-AI collaboration, particularly for generating news headlines. To tackle content creation challenges on visual platforms like Instagram Reels and TikTok, ReelFramer, a multimodal writing assistant, was developed~\cite{nickerson2023writing}. Additionally, ~\citet{cheng2024journalism} emphasized the need for customized LLMs tailored to news reporting, proposing solutions like supervised fine-tuning and constitutional AI, which integrates reinforcement learning from AI feedback to combat misinformation and rebuild reader trust. To facilitate science journalism, ~\citet{jiang2024llmcollaborationautomaticsciencejournalism} introduced a novel approach that leverages collaboration among multiple LLMs to improve the readability and clarity of news articles.

\subsection{News and Social Media Analysis} 
For news and social media analysis there has been research effort with a special focus on fact-checking \cite{shaar-etal-2022-role}, disinformation \cite{hasanain-etal-2023-araieval} and harmful content detection~\cite{lee2024trainfactverifierknowledge,alam-etal-2022-survey-1}, and news reliability classification \cite{ibrahim2024fine}. \citet{DBLP:journals/frai/QuelleB24} demonstrated that LLM agents can be employed for fact-checking by retrieving relevant evidence and verifying claims. Similarly, \citet{ibrahim2024fine} explored fine-tuned LLMs, such as Llama-3, to automate the classification of reliable versus unreliable news articles.


The \textit{Enhancing Perception}~\cite{hsu-etal-2024-enhancing} and \textit{FACT-GPT}~\cite{10.1145/3589335.3651504} frameworks tackle misinformation by refining fake news explanations through a conversational refinement approach. Similarly, VerMouth~\cite{DBLP:conf/emnlp/0004KSG23} automates social media fact-checking, contributing to broader efforts in combating misinformation. Additionally, the expert recommendation framework~\cite{10.1145/3626772.3657966} leverages a multi-layer ranking system with LLMs, balancing reliability, diversity, and comprehensiveness when suggesting experts for news events.

Other initiatives include Botlitica~\cite{musi2024botlitica}, which identifies propagandistic content in political social media posts, and JSDRV~\cite{DBLP:conf/acl/YangGM0W24}, which focuses on stance detection and rumor verification. In the realm of investigative journalism, ~\citet{ali2024enhancing} introduced a tool to retrieve and summarize relevant documents, while ~\citet{10.1145/3639701.3656308} focused on detecting framing in television program content. Addressing political bias, ~\citet{trhlik2024quantifyinggenerativemediabias} explored bias identification using LLMs. Additionally, ~\citet{10.1145/3639701.3656308} proposed prompt-engineering LLMs to analyze framing in spoken content from television programs.
A comprehensive study was conducted by \cite{zeng2024large}, highlighting the use of LLMs in social media applications.

\subsection{Specialized LLMs}
\label{ssec:specialized_llm}

\citet{ling2023domain} highlighted the importance of developing specialized models for several reasons. One key reason is that, much like humans, acquiring domain expertise and capabilities often requires years of training and experience. Therefore, it is important to train LLMs with domain knowledge to serve professional level usage.   


In this direction, a recent work by \citet{kotitsas-etal-2024-leveraging} explored fine-tuning LLMs to improve claim detection. \citet{bao2024harnessing} trained an LLM, \textit{FLLM}, using curated knowledge with a focus on the business and media domains. For training, they utilized articles published by Fortune Media. 
The OpenFactCheck framework~\cite{DBLP:journals/corr/abs-2405-05583} tackles the evaluation of factual accuracy in LLM-generated content. This customizable architecture enables the assessment of both LLM factuality and fact-checking systems, promoting standardized evaluations essential for advancing research on the reliability and factual correctness of LLMs. ~\citet{10.1145/3589334.3645471} proposed an explainable fake news detection framework that uses LLMs to generate and evaluate justifications from opposing parties. A defense-based inference module then determines veracity, improving detection accuracy and justification quality, as demonstrated on two benchmarks.

In contrast to prior work, our research focuses on developing a specialized model with a wide range of tasks and capabilities for news and social media analysis, representing the first effort in this direction to incorporate instruction-tuning and multilingual capabilities. 

\section{Tasks and Datasets}
\label{sec:task_dataset}

\subsection{Data Curation}
For dataset curation, we selected key capabilities and their associated tasks, as illustrated in Figure \ref{fig:llamalens_capablities_tasks_datasets}, and identified publicly available datasets aligned with these tasks. 
Our language choices are influenced by the demographic composition of the Gulf Cooperation Council (GCC) countries, where Arabic is predominant, English serves as a common language, and Hindi is widely spoken due to the significant Indian expatriate population.


The initial collection consists of 103 datasets, some of which we excluded due to their different versions.\footnote{For example, we have selected ArSarcasm-v2~\cite{abufarha-etal-2021-arsarcasm-v2} instead of version 1.} After initial pre-selection, the resulting collection consists of 52 datasets as detailed in Tables \ref{tab:distribution_arabic_dataset}, \ref{tab:distribution_english_dataset} and \ref{tab:distribution_hindi_dataset} in Appendix~\ref{app:sec:datasets_sizes}. 
The datasets span various sources, including social media posts, news articles, political debates and transcripts. It consists of $\sim$2.7m samples and a total of 234 labels, reflecting the complexity of tasks such as checkworthiness, claim detection, cyberbullying, emotion detection, news categorization, and more. In Appendix \ref{app:sec:datasets}, we provide detailed descriptions of the tasks and datasets.

\subsection{Preprocessing}
\label{sec:preprocss}
After collecting the datasets, we observed that while most were pre-divided into training, development, and test sets, 18 datasets lacked these splits. In cases where datasets were not pre-split, we partitioned them into 70\% for training, 20\% for testing, and 10\% for development. For datasets containing only training and test sets, we further divided the training set, allocating 30\% for development. To preserve the class distribution across splits, we employed stratified sampling.
We applied several other preprocessing steps such as removing duplicates, unifying labels (e.g., check-worthiness to checkworthiness, fixing uppercase to lowercase), and removed entries with less than 3 letters. 
These preprocessing steps ensured that the datasets were clean, well-structured, and ready for subsequent analysis or model training. 

After preprocessing, we obtained a total of 
1.2m, 1.4m and 0.14m samples for Arabic, English, and Hindi, respectively. The number of labels in the datasets ranges from 2 to 42. The datasets also include both multiclass and multilabel setups. We provide distribution of the datasets, number of labels and their splits in Tables \ref{tab:distribution_arabic_dataset}, \ref{tab:distribution_english_dataset} and \ref{tab:distribution_hindi_dataset} in Appendix~\ref{app:sec:datasets_sizes}. The datasets come in different sizes, ranging from the smallest (e.g., CT-24 subjectivity) to the largest (e.g., English news summarization dataset), and with varying label distributions, from skewed (e.g., propaganda) to more balanced (e.g., Arabic CT-22 claim detection).  
 




\section{Methodology}
\label{sec:methodology}

\begin{figure}[t]
    \centering    
    \includegraphics[scale=0.5]{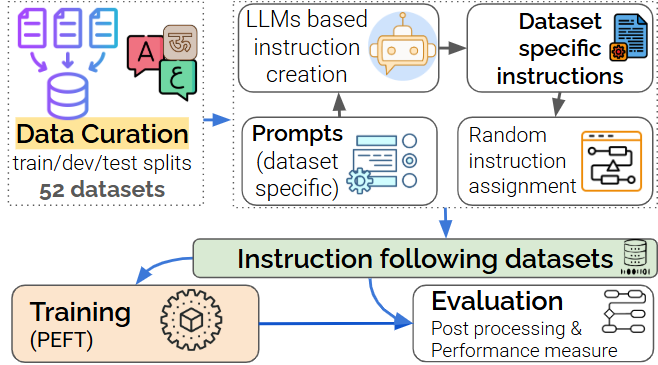}
    \caption{Approach for the \llens{}: datasets, model training, and evaluation.}
    \label{fig:methodology}
    \vspace{-0.4cm}
\end{figure}

In Figure \ref{fig:methodology}, we present the methodological steps involved in the development of \llens{}, which is also formalized in Algorithm \ref{alg1}. In Section \ref{sec:task_dataset}, we discussed the details of the dataset curation and preprocessing. The following subsections discuss the remaining steps in the development process.

\subsection{Instruction Dataset}
\label{sec:inst_dataset}
Our approach follows the typical 
pipeline of aligning LLMs with user intentions and tasks by LLM fine-tuning on representative data. Such approach usually involves creating instruction datasets starting from existing NLP datasets
~\cite{longpre2023flan}. An instruction sample is a triplet of a natural language instruction describing the task, an optional input, and an output that represents the answer following the instruction~\cite{wang2022self}.

\paragraph{Natural language instructions} There are several potential techniques to create 
natural language instructions, including manual and automatic approaches. As instructions diversity positively affects model performance and generalization~\cite{dubois2024alpacafarm,pang-etal-2024-phased,zhang2024text}, we aim to create a diverse instruction dataset. Due to the scale of tasks and datasets of focus in this work, creating such a diverse set manually is time-consuming and can lead to limited instruction styles. 

We opt to automatically generate instructions by prompting two highly-effective closed LLMs, GPT-4o
and Claude-3.5 Sonnet,
to generate a diverse set of 10 \textit{English} instructions\footnote{We chose the number 10 as a compromise between diversity and the number of samples per instruction. Finding an optimal value requires further study.} per LLM, resulting in 20 instructions per dataset. To ensure the models generate instructions fitting our datasets, we provide the models with the datasets metadata, including dataset name, language, task, task definition and labels space. Exact prompt used to generate instructions and examples of generated instructions can be found in Appendix~\ref{app:sec:app:prompting_inst}.
While findings in \cite{kmainasi2024native} show that English prompts generally outperform language-specific counterparts, we adopted a human-centric approach by providing additional native-language instructions to assess the performance of native-instructions after fine-tuning. We followed the same approach above to generate native-instructions. 

Finally, for each input dataset of the 52, we create instructions by appending a randomly selected natural instruction, of the generated 20, per example of each training subset. This guarantees versatile instruction styles even for the same input dataset. Our final instruction tuning dataset is the set of the prepared instructions for all datasets. 

\begin{algorithm}[]
\small
 \caption{Algorithm for the dataset creation and \llens{} model development. $LLM_c$ and $LLM_g$ refers to Claude 3.5 and GPT-4o models, respectively. $D'$ is the final instruction dataset. 
 \label{alg1}
 }
 \SetAlgoLined
 \KwIn{Set of datasets $\mathcal{D} = \{D_1, D_2, \dots, D_{N}\}$}
 \KwIn{$LLM_c$, $LLM_g$, $LLM_f$}
 \KwOut{Fine-Tuned Model $LLM_f'$}

 \BlankLine
 $D' \leftarrow \emptyset$\;

 \BlankLine
 \For{$k \leftarrow 1$ \KwTo $N$}{
    $I_k = \operatorname{GenerateInstruct}(LLM_c,LLM_g,D_k)$

    \BlankLine
    \ForEach{data point $d \in D_k$}{
        $i \sim \text{Uniform}(I_k)$\\
        $d' = (d, i)$ \\
        $D' \leftarrow D' \cup \{d'\}$
    }
 }
 \BlankLine
 Fine-tune the model $LLM_f$ using $D'$:
 $LLM_f' = \operatorname{FineTune}(LLM_f, D')$
 \BlankLine
 \KwRet{Fine-Tuned Model $LLM_f'$}\;
\end{algorithm}


\subsection{Training}
We base our experiments on Llama 3.1, the most effective \textit{open} LLM to date, with strong multilingual performance~\cite{dubey2024llama}. Fine-tuning larger scales of the model (e.g., 70B version) holds a great overhead in terms of time and computational cost. Moreover, These models may be inaccessible to many researchers, so we focus on the smaller Llama 3.1-8B version. We base \llens{} on Llama 3.1-8B-Instruct, as it is already aligned with various user tasks.

\subsubsection{Training Setups}
We instruction-tune Llama 3.1-8B-Instruct\footnote{We use the term \textit{Llama-instruct} or \textit{Base} to refer to this model in the rest of the paper.} following different setups. Given that fine-tuning LLMs typically requires substantial computational resources, making it a time-consuming and resource-intensive process, therefore, to address this challenge, our main \llens{} model is fine-tuned in bf16 following parameter-efficient fine-tuning using Low-Rank Adaptation (LoRA)~\cite{hu2021lora}. 
In addition to the full precision model, we aimed to train smaller models, achieving two goals: \textit{(i)} release smaller but effective models to be used in resource-constrained environments, and \textit{(ii)} efficiently investigate the effects of some parameter settings on model performance, to guide the full model training. Thus, we also fine-tune the original Llama-8B-Instruct model employing QLoRA~\cite{dettmers2024qlora}, which involves quantization of the model's weights and significantly enhances memory optimization, 
 while maintaining acceptable performance.

\begin{table*}[t]
\setlength{\tabcolsep}{2.5pt}
\centering
\resizebox{\textwidth}{!}{%
\begin{tabular}{lll|llllr|lllllllr}
\toprule
\textbf{Task}   & \textbf{Dataset}   & \textbf{Metric}  & {\textbf{SOTA}}  &\textbf{Base}& \textbf{\shortstack{L-Lens \\ (Nat.)}} & \textbf{\shortstack{L-Lens \\ (Eng.)}} & \multicolumn{1}{c}{\textbf{$\Delta$}} & {\textbf{Task}}  & {\textbf{Dataset}}  & {\textbf{Metric}}  & {\textbf{SOTA}}  &\textbf{Base}& \textbf{\shortstack{L-Lens \\ (Nat.)}} & \textbf{\shortstack{L-Lens \\ (Eng.)}} & \multicolumn{1}{c}{\textbf{$\Delta$}} \\
\midrule
\multicolumn{8}{c|}{\textbf{Arabic}} & & \multicolumn{6}{c}{\textbf{English}} & \\ 
\midrule

Attention. & CT22Attentionworthy   & W-F1    & 0.412   &0.158
& \textbf{0.454}& 0.425& \underline{0.013}& Check. &CT24\_T1   & F1\_Pos  & 0.753   &0.404
& 0.942& \textbf{0.942}& \underline{0.189}\\
Check. &CT24\_T1 & F1\_Pos & 0.569&\textbf{0.610}& 0.509& 0.502
& -0.067
& Claim  & claim-detection & Mi-F1 & \multicolumn{1}{c}{--} &  0.545
& \textbf{0.889}& 0.864
& \multicolumn{1}{c}{--}   \\
Claim  & CT22Claim & Acc & 0.703   &0.581
& \textbf{0.756}& 0.734& \underline{0.031}& Cyberbullying & Cyberbullying  & Acc    & \textbf{0.907$^*$} &0.175
& 0.855& 0.836& -0.071
\\
Cyberbullying  & ArCyc\_CB & Acc & 0.863$^*$   &0.766
& 0.833& \textbf{0.870}& \underline{0.007}& Emotion  & emotion   & Ma-F1    & 0.790$^*$   &0.353
& \textbf{0.808}& 0.803
& \underline{0.013}\\
Emotion & Emotional-Tone & W-F1    & 0.658$^*$   &0.358
& \textbf{0.736}& 0.705
& \underline{0.047}& Factuality    & News\_dataset  & Acc & 0.920$^*$   &0.654
& 0.999& \textbf{0.999}& \underline{0.080}\\
Emotion & NewsHeadline   & Acc & \textbf{1.000$^*$} &0.406
& 0.458& 0.480& -0.520
& Factuality    & Politifact & W-F1 & \textbf{0.490$^*$} &0.121
& 0.311& 0.287
& -0.203
\\
Factuality   & Arafacts  & Mi-F1   & \textbf{0.850$^*$} &0.210
& 0.738& 0.771& -0.079
& News Cat. & CNN\_News\_Articles  & Acc    & 0.940   &0.644
& 0.970& \textbf{0.970}& \underline{0.030}\\
Factuality   & COVID19Factuality & W-F1    & 0.838   &0.492
& \textbf{0.840}& 0.800& -0.031
& News Cat. & News\_Category  & Ma-F1    & 0.769$^*$   &\textbf{0.970}& 0.520& 0.824& \underline{0.055}\\
Harmfulness  & CT22Harmful    & F1\_Pos & \textbf{0.557} &0.507
& 0.535& 0.523
& -0.034
& News Genre    & SemEval23T3-ST1 & Mi-F1    & \textbf{0.815} &0.687
& 0.253& 0.241
& -0.574
\\
Hate Speech  & annotated-hatetweets-4 & W-F1    & \textbf{0.630} &0.257
& 0.517& 0.526& -0.104
& News Sum.  & xlsum & R-2 & 0.152   &0.074
& 0.181& \textbf{0.182}& \underline{0.030}\\
Hate Speech  & OSACT4SubtaskB & Mi-F1   & 0.950   &0.819
& \textbf{0.955}& 0.955
& \underline{0.005}& Offensive  & Offensive\_Hateful & Mi-F1 & \multicolumn{1}{c}{--} &  0.692
& 0.813& \textbf{0.814}& \multicolumn{1}{c}{--}   \\
News Cat. & ASND & Ma-F1   & 0.770$^*$   &0.587
& \textbf{0.929}& 0.919
& \underline{0.149}& Offensive  & offensive\_language    & Mi-F1 & \textbf{0.994} &0.646
& 0.893& 0.899
& -0.095
\\
News Cat. & SANADAkhbarona & Acc & 0.940   &0.784
& 0.953& \textbf{0.954}& \underline{0.014}& Offensive \& Hate  & hate-offensive-speech    & Acc & \textbf{0.945} &0.602
& 0.935& 0.931
& -0.014
\\
News Cat. & SANADAlArabiya & Acc & \textbf{0.974} &0.893
& 0.985& 0.987
& \underline{0.013}& Propaganda    & QProp & Ma-F1 & 0.667   &0.759
& \textbf{0.973}& 0.963
& \underline{0.296}\\
News Cat. & SANADAlkhaleej & Acc & 0.968   &0.865
& 0.982& \textbf{0.984}& -0.002
& Sarcasm  & News-Headlines & Acc & 0.897$^*$   &0.668
& \textbf{0.947}& 0.936
& \underline{0.039}\\
News Cat. & UltimateDataset  & Ma-F1   & \textbf{0.970} &0.376& 0.880& 0.865& -0.105
& Sentiment & NewsMTSC    & Ma-F1    & \textbf{0.817} &0.628
& 0.748& 0.751& -0.066
\\
News Cred.    & NewsCredibility & Acc & 0.899$^*$   &0.455
& 0.933& \textbf{0.935}& \underline{0.036}& Subjectivity  & CT24\_T2   & Ma-F1 & \textbf{0.744} &0.535& 0.628& 0.642& -0.102\\

\cmidrule{9-16}

News Sum.  & xlsum & R-2 & \textbf{0.137} &0.034
& 0.130& 0.129
& -0.009 &

\multicolumn{3}{c}{\textbf{\textit{Average}}} & \textbf{0.773} & 0.528 & 0.731  & 0.747 & -0.026   \\

\cmidrule{9-16}

Offensive  & OSACT4SubtaskA & Ma-F1   & \textbf{0.905} &0.782
& 0.882& 0.896
& -0.009

& \multicolumn{7}{c}{\textbf{Hindi}} &  \\
\cmidrule{9-16}
Offensive   & ArCyc\_OFF & Ma-F1   & 0.878$^*$   &0.489
& \textbf{0.879}& 0.877
& -0.001

& Factuality    & fake-news & Mi-F1 & \multicolumn{1}{c}{--} &  0.759
& 0.993& \textbf{0.994}& \multicolumn{1}{c}{--}   \\

Propaganda   & ArPro & Mi-F1   & \textbf{0.767} &0.597
& 0.731& 0.747
& -0.020
& Hate Speech   & hate-speech-detection    & Mi-F1    & 0.639$^*$   &0.750
& 0.963& \textbf{0.963}& \underline{0.324}\\

Sarcasm & ArSarcasm-v2   & F1\_Pos & \textbf{0.584} &0.477
& 0.542& 0.520
& -0.064
& Hate Speech   & Hindi-Hostility    & W-F1 & \textbf{0.841$^*$} &0.469
& 0.753& 0.753
& -0.088
\\

Sentiment    & ar\_reviews\_100k & F1\_Pos & \multicolumn{1}{c}{--} &  0.681
& 0.779& \textbf{0.785}& {--}   
& NLI & NLI\_dataset & W-F1 & 0.646   &0.633
& \textbf{0.679}& 0.568
& -0.078
\\

Sentiment    & ArSAS & Acc & \textbf{0.930$^*$} &0.603& 0.804& 0.800
& -0.120
& News Sum.  & xlsum & R-2 & 0.136   &0.078
& 0.170& \textbf{0.171}& \underline{0.035}\\

Stance & stance    & Ma-F1   & 0.767   &0.608
& 0.881& \textbf{0.926}& \underline{0.159}
& Offensive  & Offensive Speech & Mi-F1    & 0.723   &0.621
& \textbf{0.865}& 0.862
& \underline{0.139}\\

Stance & Mawqif-Arabic & Ma \-f1 & 0.789 & 0.764 & 0.826 & \textbf{0.853} & \underline{0.065}
& Cyberbullying & MC-Hinglish1.0  & Acc & 0.609   &0.233
& \textbf{0.627}& 0.625
& \underline{0.016} \\

 Subjectivity & ThatiAR   & F1\_Pos & \textbf{0.800} &0.562& 0.383& 0.441& -0.359
 &
Sentiment & Sentiment Analysis  & Acc & \textbf{0.697} &0.552
& 0.654& 0.647& -0.050\\

\midrule 

\multicolumn{2}{c}{\textbf{\textit{Average}}} & ~ & \textbf{0.773} & 0.540 & 0.733 & 0.735 & -0.038 & \multicolumn{3}{c}{\textbf{\textit{Average}}} & 0.613 & 0.477  & \textbf{0.673}  & 0.656 & \underline{0.043} \\

\bottomrule

\end{tabular}%
}
\vspace{-0.2cm}
\caption{\llens{} performance across all datasets. SOTA: State-of-the-art results. L-Lens: \textbf{LlamaLens}, Nat.: Native, Eng.: English. Base:\textbf{ Llama 3.1-8B-Instruct}. R-2: ROUGE-2, Acc: Accuracy, Mi-F1: Micro-averaged F1, Ma-F1: Macro-averaged F1, W-F1: Weighted F1, F1\_Pos: F1 score for the positive class, --: No SOTA scores found. NLI: Natural Language Inference. Attention: Attentionworthiness. News Cred: News Credibility. News Sum.: News Summarization. Cat.: Categorization. Check: Checkworthiness   $^*$: Data was not pre-split. The $\Delta$ column represents the difference between \llens{} (Eng.) and the corresponding SOTA value. \textbf{Underlined} values in the $\Delta$ column indicate cases where \llens{} (Eng.) outperforms the SOTA.}
\label{tab:fullresults}
\vspace{-0.35cm}
\end{table*}

\subsubsection{Experimental Setup}
\paragraph{Datasets Sampling}
Our experimental dataset covers 52 distinct datasets. As explained in Section~\ref{sec:inst_dataset}, we aimed to create a diverse instruction dataset starting from these NLP datasets. Given the substantial size of some of these datasets, for example, the Arabic hate speech dataset contains 0.2$M$ samples, and to ensure versatility, we pragmatically set a threshold of 20$K$ training instances per dataset. For datasets exceeding this limit, we employed stratified sampling to preserve the original distribution of the dataset labels. Our final training dataset includes 0.6$M$ samples out of 1.96$M$. 
We will release the complete training set of instructions for future studies. 

\paragraph{Dataset Shuffling}
The order of instructions in the training dataset can significantly impact model performance. For example, \citet{pang-etal-2024-phased} demonstrate that ordering instructions by complexity influences the effectiveness of tuned models. In light of this, we investigate how different orders of samples affect the performance of \llens{}, employing four different data shuffling and ordering techniques to identify the optimal sequence. Although multiple ordering configurations exist, we focus 
on the effects of language and task order.

\begin{enumerate} [noitemsep,topsep=0pt,leftmargin=*,labelsep=.5em]
    \item \textbf{Alphabetically ordered}: This is a basic setup where languages and datasets are ordered alphabetically—Arabic, followed by English, and then Hindi—without shuffling.
    \item \textbf{Shuffled by language}: Randomly shuffle datasets while preserving the order of languages.
    \item \textbf{Shuffled by task}: Tasks are organized alphabetically, with datasets shuffled across tasks regardless of language.
    \item \textbf{Fully randomized}: Complete randomization of the training dataset points.
\end{enumerate}




\paragraph{Parameters Setup}

For all models we train, we set a LoRA learning rate of 2e-4. Optimization was performed using AdamW \cite{loshchilov2017decoupled}, with a batch size of 16. All experiments were executed on four NVIDIA H100-80GB GPUs.

In the first set of experiments, we trained four models quantized to 4-bit precision using QLoRA. Although the models store weights in 4-bit format, computations are performed in BFLOAT16 (bf16), with both the LoRA rank and $\alpha$ set to 16. Each model was trained on one of the dataset shuffling configurations. After identifying the optimal order of the training dataset—based on average model performance on our test sets, as shown in Figure~\ref{fig:quantized_performance}—we used that dataset order to fine-tune our \llens{} model in full precision (16-bit). Due to the scale of the model and training set size, we train the model for two epochs, increasing LoRA rank to 128, following the recommendations in \cite{xin-etal-2024-beyond}, which suggests that higher ranks yield improved performance for multitask learning. LoRA $\alpha$ was set equal to the rank. 

\subsection{Evaluation}
For the evaluation, we employed a zero-shot approach, in which we directly prompted the models 
to perform tasks from the testing sets. The employed natural instruction/prompt is the first generated instruction (Section~\ref{sec:inst_dataset}) the \textit{temperature} was set to 0 and \textit{top\_p} to 0.95. We manually checked a sample of instructions per task and found that they accurately expressed the intended task.\footnote{Scripts are available at: \url{https://github.com/firojalam/LlamaLens}} 


\subsection{Post-processing}
As models can generate text beyond that is required in the instruction, a post-processing method was implemented to extract labels from the generated models' responses. Initially, a regular expression was used to accurately identify and extract the labels. 
Several transformations were applied, including lowercasing, removing special characters, and handling code-switching by replacing non-Latin characters with Latin equivalents, similar to \citet{abdelali-etal-2024-larabench,dalvi-etal-2024-llmebench}.


\subsection{Evaluation Metrics}
All models were evaluated using standard classification metrics: weighted, macro, micro F1, and accuracy. Summarization was assessed with ROUGE-2,  Specifically, we use the same metric reported in state-of-the-art (SOTA) per dataset.

\begin{table*}[t]
\centering
\setlength{\tabcolsep}{2.3pt}
\scalebox{0.73}{
\begin{tabular}{@{}llrrrrr|llrrrrr@{}}
\toprule
\multirow{2}{*}{\textbf{Dataset}} & \multirow{2}{*}{\textbf{Metric}} & \multicolumn{5}{c|}{\textbf{Model Performance}} & \multirow{2}{*}{\textbf{Dataset}} & \multirow{2}{*}{\textbf{Metric}} & \multicolumn{5}{c}{\textbf{Model Performance}} \\ 
\cmidrule(lr){3-7} \cmidrule(lr){10-14} 
& & \textbf{Base} & \textbf{Alpha.} & \textbf{Full} & \textbf{Task} & \textbf{Lang.} & & & \textbf{Base} & \textbf{Alpha.} & \textbf{Full} & \textbf{Task} & \textbf{Lang.} \\ 
\midrule
\multicolumn{7}{c|}{\textbf{Arabic}} & \multicolumn{7}{c}{\textbf{English}} \\ 
\midrule

CT22Attentionworthy & W-F1 & 0.158 & 0.281 & 0.299 & 0.293 & \textbf{0.340} & 

CT24\_T1 & F1\_Pos & 0.404 & 0.538 & 0.583 & \textbf{0.893} & 0.288 \\

CT24\_T1 & F1\_Pos & 0.610 & 0.416 & 0.555 & \textbf{0.689} & 0.549  &

claim-detection & Mi-F1 & 0.545 & 0.895 & 0.891 & 0.884 & \textbf{0.898} \\

CT22Claim & Acc & 0.581 & 0.712 & \textbf{0.735} & 0.715 & 0.723 & Cyberbullying & Acc & 0.175 & 0.664 & \textbf{0.794} & 0.781 & 0.764 \\

 ArCyc\_CB & Acc & 0.766 & 0.767 & 0.818 & \textbf{0.840} & 0.776 & emotion & Ma-F1 & 0.353 & 0.647 & \textbf{0.662} & 0.584 & 0.654 \\

 Emotional-Tone & W-F1 & 0.358 & 0.595 & 0.635 & \textbf{0.641} & 0.609 & News\_dataset & Acc & 0.654 & 0.502 & 0.712 & \textbf{0.787} & 0.614 \\

NewsHeadline & Acc & \textbf{0.406} & 0.316 & 0.319 & 0.387 & 0.288 & Politifact & W-F1 & 0.121 & 0.210 & 0.241 & 0.252 & \textbf{0.262} \\

Arafacts & Mi-F1 & 0.210 & 0.376 & 0.263 & \textbf{0.466} & 0.362 & CNN\_News\_Articles & Acc & 0.644 & 0.897 & \textbf{0.919} & 0.904 & 0.911 \\

COVID19Factuality & W-F1 & 0.492 & \textbf{0.794} & 0.733 & 0.595 & 0.780 & News\_Category & Ma-F1 & \textbf{0.970} & 0.964 & 0.913 & 0.635 & 0.668 \\

 CT22Harmful & F1\_pos & 0.507 & 0.539 & \textbf{0.565} & 0.473 & 0.539 & SemEval23T3-ST1 & Mi-F1 & \textbf{0.687} & 0.325 & 0.494 & 0.470 & 0.590 \\

 annotated-hatetweets-4 & W-F1 & 0.257 & \textbf{0.436} & 0.311 & 0.371 & 0.394 & xlsum & R-2 & 0.074 & 0.088 & \textbf{0.126} & 0.116 & 0.101 \\

OSACT4SubtaskB & Mi-F1 & 0.819 & \textbf{0.946} & 0.901 & 0.910 & 0.911 & Offensive\_Hateful & Mi-F1 & 0.692 & 0.791 & 0.768 & \textbf{0.792} & 0.778 \\

ASND & Ma-F1 & 0.587 & 0.790 & 0.787 & 0.803 & \textbf{0.811} & offensive\_language & Mi-F1 & 0.646 & \textbf{0.893} & 0.871 & 0.657 & 0.821 \\

SANADAkhbarona & Acc & 0.784 & 0.924 & 0.904 & 0.930 & \textbf{0.938} & hate-offensive-speech & Acc & 0.602 & 0.874 & 0.901 & \textbf{0.909} & 0.903 \\

 SANADAlArabiya & Acc & 0.893 & \textbf{0.975} & 0.973 & 0.973 & 0.973 & QProp & Ma-F1 & 0.759 & 0.773 & \textbf{0.803} & 0.751 & 0.773 \\

SANADAlkhaleej & Acc & 0.865 & 0.929 & 0.920 & 0.916 & \textbf{0.929} & News-Headlines & Acc & 0.668 & 0.959 & \textbf{0.961} & 0.953 & 0.960 \\

UltimateDataset & Ma-F1 & 0.376 & \textbf{0.742} & 0.594 & 0.647 & 0.673 & NewsMTSC-dataset & Ma-F1 & 0.628 & 0.640 & 0.669 & \textbf{0.685} & 0.613 \\

NewsCredibility & Acc & 0.455 & 0.665 & 0.845 & \textbf{0.904} & 0.600 & CT24\_T2 & Ma-F1 & 0.535 & 0.464 & 0.440 & \textbf{0.554} & 0.379 \\

\cmidrule{8-14}
 ArCyc\_OFF & Ma-F1 & 0.489 & 0.835 & 0.846 & \textbf{0.856} & 0.836 & 
\multicolumn{2}{c} {\textbf{\textit{Average}}} & 0.528  & 0.629  & \textbf{0.673} & 0.662 &  0.620  \\
\cmidrule{8-14}

xlsum & R-2 & 0.034 & 0.058 & 0.058 & 0.061 & \textbf{0.063} &
\multicolumn{7}{c}{\textbf{Hindi}} \\  

\cmidrule{8-14}

 OSACT4SubtaskA & Ma-F1 & 0.782 & \textbf{0.876} & 0.852 & 0.863 & 0.849 & fake-news & Mi-F1 & 0.759 & \textbf{0.802} & 0.633 & 0.567 & 0.653 \\

ArPro & Mi-F1 & 0.597 & 0.660 & 0.623 & \textbf{0.696} & 0.655 & hate-speech-detection & Mi-F1 & 0.750 & \textbf{0.910} & 0.898 & 0.903 & 0.879 \\

 ArSarcasm-v2 & F1\_Pos & 0.477 & 0.154 & \textbf{0.542} & 0.429 & 0.472 & Hindi-Hostility & W-F1 & 0.469 & \textbf{0.666} & 0.564 & 0.664 & 0.526 \\

ar\_reviews\_100k & F1\_Pos & \textbf{0.681} & 0.552 & 0.626 & 0.614 & 0.626 & NLI\_dataset & W-F1 & \textbf{0.633} & 0.516 & 0.573 & 0.537 & 0.564 \\

ArSAS & Acc & 0.603 & \textbf{0.780} & 0.774 & 0.763 & 0.772  & xlsum & R-2 & 0.078 & 0.074 & 0.094 & \textbf{0.095} & 0.080 \\

 stance & Ma-F1 & 0.608 & 0.634 & 0.774 & 0.775 & \textbf{0.853}& Offensive\_Speech & Mi-F1 & 0.621 & 0.692 & 0.701 & 0.733 & \textbf{0.763} \\

Mawqif-Arabic-Stance & Ma-F1 & 0.764 & 0.774 & 0.819 & 0.845 & \textbf{0.846}  & MC-Hinglish1.0 & Acc & 0.233 & 0.640 & \textbf{0.643} & 0.636 & 0.545 \\

ThatiAR & F1\_Pos & 0.562 & 0.558 & 0.544 & 0.574 & \textbf{0.591} 
& Sentiment Analysis & Acc & 0.552 & 0.627 & 0.650 & \textbf{0.658} & 0.624 \\

\midrule

\multicolumn{2}{c}{\textbf{\textit{Average}}} & 0.540 & 0.636 & 0.653 & \textbf{0.668} & 0.659 &

\multicolumn{2}{c}{\textbf{\textit{Average}}} & 0.4777 & 0.589 & 0.589 & \textbf{0.604} &  0.569 \\
 \bottomrule
\end{tabular}
}
\vspace{-0.2cm}
\caption{Performance of models trained with QLoRA across all datasets and dataset shuffling techniques. R-2: ROUGE-2.  Acc: Accuracy. Mi-F1: Micro-averaged F1.  Ma-F1: Macro-averaged F1. W-F1: weighted F1. F1\_Pos: F1 of the positive class.
\textbf{Base}: Llama 3.1-8B-Instruct.  
\textbf{Alpha}: Model trained with QLoRA on a dataset ordered alphabetically.  
\textbf{Full}: Model trained with QLoRA on a fully randomized dataset.  
\textbf{Task}: Model trained with QLoRA on a dataset shuffled by task.  
\textbf{Lang.}: Model trained with QLoRA on a dataset shuffled by language.  
Numbers in \textbf{bold} indicate the best performance per dataset.
}
\label{tab:quant_perform}
\vspace{-0.3cm}
\end{table*}

\section{Results and Discussions}
\label{sec:results_and_discussions}

To contextualize the performance of \llens{}, we compare it against both the SOTA results and the \textit{Llama-instruct} 
baseline. SOTA serves as a theoretical upper bound, representing the best-reported results on each dataset, while the \textit{Llama-instruct} model acts as a lower bound, reflecting a general-purpose instruction-tuned LLM without task-specific adaptation. Our evaluation aims to assess how well \llens{} bridges the gap between these two extremes, demonstrating its ability to generalize while leveraging task-specific fine-tuning.





\subsection{\llens{} Performance}

In Table \ref{tab:fullresults}, we report the full results of our \llens{} model across the different languages. Overall, \llens{} significantly outperforms the Llama-instruct 
with an average improvement of 62\%. 
Language wise \llens{} outperforms Llama-instruct on average 59\%, 71\% and 56\% for Arabic, English and Hindi, respectively. 
For Arabic, \llens{} demonstrates strong performance across 27 tasks, underperforming only in \textit{Checkworthiness} and one \textit{Subjectivity} dataset. In English, \llens{} outperforms in 15 out of 17 tasks, with \textit{News Categorization} being the primary area of lower performance. For Hindi, across 8 datasets, \llens{} surpasses Llama-Instruct in 7, with \textit{Natural Language Inference} being the only dataset where it under-performs compared to Llama-Instruct. 

\noindent
\textbf{Comparable Performance with SOTA.} In comparison to SOTA's average performance of 0.75, \llens{} achieves an average performance of 0.727. We should note that since 18 out of the 52 datasets were not pre-split (Section~\ref{sec:preprocss}), SOTA on these datasets is not directly comparable to our model, as the testing splits might differ. For some datasets, issues such as duplicate entries, missing input text, or the absence of development or test splits (e.g., \textit{CNN\_News\_Articles} for English) required us to clean the dataset and create new splits. As a result, a direct apple-to-apple comparison may not always be possible. However, the reported SOTA scores serve as a close approximation for meaningful evaluation. Computing the average performance excluding these datasets reduces the gap, with a SOTA of 0.716 and \llens{} performance of 0.693.
In terms of number of datasets where \llens{} improved over SOTA, we find that it outperforms SOTA in 23 test sets, and has a comparable performance (difference between -0.04 to 0)
in 8 other testing sets.


\noindent
\textbf{Performance Gains Across Datasets.} Dataset-specific improvements in English include gains in \textit{Factuality}, with \textit{Propaganda} and \textit{Checkworthiness} showing significant advancements. \textit{Summarization} improved across both English and Hindi, along with \textit{News Genre Categorization} (two datasets in English and four in Arabic). In Hindi, \textit{Offensive Language} demonstrated notable improvements, while \textit{Hate Speech} and \textit{Cyberbullying} also exhibited gains, with the latter performing better in Arabic as well. Additionally, Arabic showed stronger performance in \textit{News Credibility}, \textit{Emotion}, \textit{Stance}, \textit{Attentionworthiness}, and \textit{Claim Detection}.


\subsection{Native vs. English Instructions}
When comparing the two versions of \llens{} (trained with English vs. Native instructions), the performance difference between them is minimal. The results are closely aligned, with the Native model\footnote{The model was trained using language-specific instructions tailored to the dataset.} achieving an average score of 0.723, compared to 0.727 for the English-instructed model. 
A closer look reveals that the English-instructed model outperformed the Native model in 28 datasets, while the Native model led in 24. Overall, 49 of 52 datasets showed comparable performance, with differences between -0.05 and 0.05. The most notable difference was in \textit{News\_Category\_Dataset}, where the English-instructed model outperformed the Native model by 0.304.

\subsection{Impact of Data Shuffling} 
\begin{figure}[h]
    \centering
    \includegraphics[width=1\linewidth]{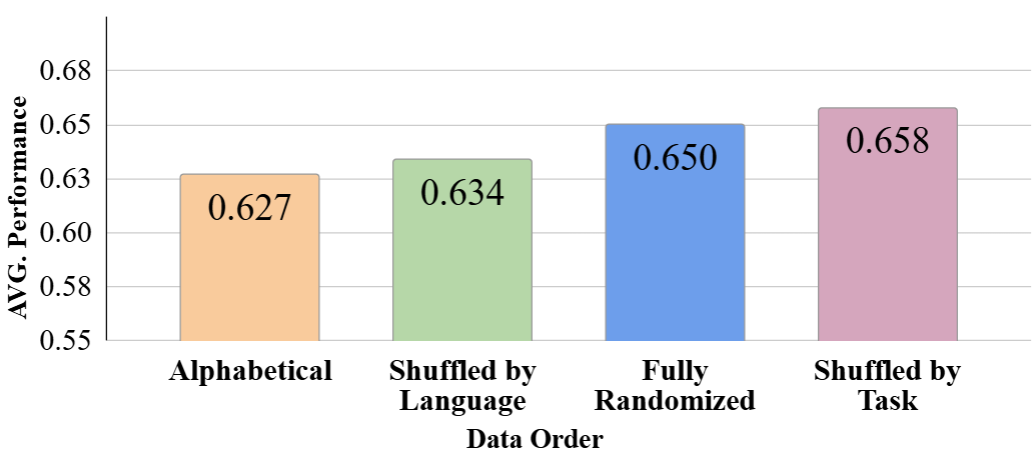}
    \vspace{-0.5cm}
    \caption{Impact of data shuffling technique on fine-tuned quantized Llama-3.1 performance.}
    \label{fig:quantized_performance}
    \vspace{-0.5    cm}
\end{figure}

Figure \ref{fig:quantized_performance} shows the averaged results across datasets and languages, with details in Table \ref{tab:quant_perform}. \textit{Shuffling by task} achieved the highest performance, while \textit{shuffling by language} and \textit{alphabetic ordering} performed similarly but did not match the effectiveness of the task-based approach.
To determine statistical significance, we performed a Wilcoxon signed-rank test comparing the best (shuffled by task) and worst (alphabetical) configurations. The test yielded a p-value of 0.025, below the \(\alpha = 0.05\) threshold, confirming that improvements from task-based shuffling are not random. Based on this, we adopt \textit{shuffling by task} for training \llens{} to enhance performance across diverse datasets. 

\subsection{Task-wise Results}
We computed task-wise performance differences, as detailed in Table \ref{Task-Capability-Based}. Fine-tuned models demonstrated significant improvements, particularly in English Cyberbullying tasks, where \llens{} achieved a performance gain of more than 61\% compared to other languages. Across all three languages, \llens{} consistently outperformed baselines in \textit{sentiment analysis}, \textit{summarization}, \textit{factuality}, \textit{hate speech}, and \textit{offensive language detection}. Notably, when evaluated across 37 combined datasets, \llens{} underperformed in only four cases: \textit{natural language inference}, \textit{subjectivity (Arabic)}, \textit{checkworthiness (Arabic)}, and \textit{news genre categorization} (specifically the SemEval23T3-ST1 dataset in English).

\begin{table}[!ht]
\centering

\setlength{\tabcolsep}{4pt}
\scalebox{0.73}{
\begin{tabular}{@{}llrccc@{}}
    \toprule
    \textbf{Task} & \textbf{Lang} & \textbf{\#DS} 
    & \textbf{Base} & \textbf{\shortstack{L-Lens \\ (Eng.)}} & \textbf{$\Delta$} \\
    \midrule
    \multirow{3}{*}{\textbf{Summarization}} 
    & Arabic  & 1 & 0.034 & 0.129 & 0.095 \\
    & English & 1 & 0.074 & 0.182 & 0.108 \\
    & Hindi   & 1 & 0.078 & 0.171 & 0.093 \\
    \midrule
    \multirow{2}{*}{\textbf{News Cat.}} 
    & Arabic  & 5 & 0.701 & 0.942 & 0.241 \\
    & English & 3 & 0.767 & 0.678 & -0.088 \\
    \midrule
    \multirow{2}{*}{\textbf{Emotion}} 
    & Arabic  & 2 & 0.382 & 0.592 & 0.211 \\
    & English & 1 & 0.353 & 0.803 & 0.450 \\
    \midrule
    \multirow{2}{*}{\textbf{Sarcasm}} 
    & Arabic  & 1 & 0.477 & 0.520 & 0.043 \\
    & English & 1 & 0.668 & 0.936 & 0.268 \\
    \midrule
    \multirow{3}{*}{\textbf{Sentiment}} 
    & Arabic  & 2 & 0.642 & 0.792 & 0.150 \\
    & English & 1 & 0.628 & 0.751 & 0.123 \\
    & Hindi   & 1 & 0.552 & 0.647 & 0.095 \\
    \midrule
    \textbf{Stance} & Arabic & 2 & 0.686 & 0.890 & 0.203 \\
    \textbf{News Credibility} & Arabic & 1 & 0.455 & 0.935 & 0.480 \\
    \textbf{Attentionworthy} & Arabic & 1 & 0.158 & 0.425 & 0.267 \\
    \midrule
    \multirow{2}{*}{\textbf{Checkworthiness}} 
    & Arabic  & 1 & 0.610 & 0.425 & -0.185\\
    & English & 1 & 0.404 & 0.942 & 0.539 \\
    \midrule
    \multirow{2}{*}{\textbf{Claim}} 
    & Arabic  & 1 & 0.581 & 0.734 & 0.153 \\
    & English & 1 & 0.545 & 0.864 & 0.319 \\
    \midrule
    \multirow{3}{*}{\textbf{Factuality}} 
    & Arabic  & 2 & 0.351 & 0.785 & 0.434 \\
    & English & 2 & 0.387 & 0.643 & 0.256 \\
    & Hindi   & 1 & 0.759 & 0.994 & 0.235 \\
    \midrule
    \multirow{2}{*}{\textbf{Propaganda}} 
    & Arabic  & 1 & 0.597 & 0.747 & 0.150 \\
    & English & 1 & 0.759 & 0.963 & 0.204 \\
    \midrule
    \multirow{3}{*}{\textbf{Cyberbullying}} 
    & Arabic  & 1 & 0.766 & 0.870 & 0.104 \\
    & English & 1 & 0.175 & 0.836 & 0.661 \\
    & Hindi   & 1 & 0.233 & 0.625 & 0.392 \\
    \midrule
    \textbf{Harmfulness} & Arabic & 1 & 0.507 & 0.523 & 0.016 \\
    \midrule
    \multirow{3}{*}{\textbf{Hate Speech}} 
    & Arabic  & 2 & 0.538 & 0.740 & 0.203 \\
    & English & 1 & 0.602 & 0.931 & 0.329 \\
    & Hindi   & 2 & 0.609 & 0.858 & 0.249 \\
    \midrule
    \multirow{3}{*}{\textbf{Offensive}} 
    & Arabic  & 2 & 0.636 & 0.887 & 0.251 \\
    & English & 3 & 0.647 & 0.881 & 0.235 \\
    & Hindi   & 1 & 0.621 & 0.862 & 0.241 \\
    \midrule
    \multirow{2}{*}{\textbf{Subjectivity}} 
    & English & 1 & 0.535 & 0.642 & 0.107 \\
    & Arabic  & 1 & 0.562 & 0.441 & -0.121 \\
    \midrule
    \textbf{NLI} & Hindi & 1 & 0.633 & 0.568 & -0.065 \\
    \bottomrule
\end{tabular}
}
\vspace{-0.2cm}
\caption{Task-Based Evaluation Across Arabic, English, and Hindi. Lang: Languages. \#DS: Number of Datasets. Base: Llama3.1-Instruct. L-Lens: LlamaLens. Eng: English. News Cat: News Genre Categorization. NLI: Natural Language Inference. The \textbf{$\Delta$} Column represents the difference between two models. 
}
\label{Task-Capability-Based}
\vspace{-0.35cm}
\end{table}


\subsection{Capability-based Results}

Table \ref{Arabic-Capability-Based} presents LlamaLens's performance across key task categories for Arabic, while Table \ref{English-Hindi-Capability-Based} covers English and Hindi. These categories include \textit{Natural Language Understanding (NLU)}, \textit{Information Extraction} and \textit{Text Classification}, \textit{Emotion and Sentiment Analysis}, \textit{Fact-Checking and Verification}, and \textit{Hate Speech and Offensive Content}, as illustrated in Figure \ref{fig:llamalens_capablities_tasks_datasets}. The tables report the SOTA scores, the Llama-instruct baseline, and \llens{} using native instructions.

\llens{} achieves notable gains over Llama-instruct, particularly in Arabic and English, with English tasks consistently performing better due to the availability of abundant datasets and established benchmarks. Despite being a medium-resource language, Arabic shows significant improvements, closing the performance gap with English across multiple tasks. For Hindi, while overall scores remain lower, \llens{} demonstrates clear advancements, particularly in \textit{emotion and sentiment analysis} and \textit{hate speech} and \textit{offensive content}, highlighting its potential in underrepresented languages. Hate speech and offensive Content emerges as \llens{}'s strongest task across all three languages, with the highest improvement recorded in English (+0.411), demonstrating its capability in handling complex linguistic challenges.

\subsection{Error Analysis} 
We analyzed Llama-Instruct’s responses to identify challenges across different tasks, such as handling offensive language, hate speech, factuality, and news categorization.

\noindent
\textbf{Model Hesitation and Contextual Dependence.} One recurring problem was the inability of the model to provide labels in numerous instances, often responding with phrases like \textit{``I cannot provide a label''} or \textit{``Arabic text is not easily classifiable into categories without context or translation.''} Such responses occur when an LLM lacks context to classify text confidently or is designed to avoid labeling sensitive topics (e.g., political, religious, or controversial). These hesitations stem from built-in safeguards to prevent incorrect classifications.

\noindent
\textbf{Language Confusion.} Another observed issue is language confusion in output. Although the models were instructed to output labels exclusively in English, they occasionally returned labels in Arabic or code-switched responses, which is inline with language confusion reported in relevant studies~\cite{marchisio2024understanding}, however, differently, we showcase the confusion can occur at the smallest unit of a single character. For instance, in some cases, the model generated outputs like ``\<ف>actual'' (referring to ``factual'' where Arabic character \<ف> is a transliteration of character ``f''), and ``\<س>포츠'' (Korean for ``sport'' transliterated as ``seupocheu'' where character \<س> is a transliteration of character ``s'') despite no instructions involving Korean language. This highlights a phonetic-level code-switching phenomenon. It also occurred on longer sequences such as the model responding with ``\<سار>castic'' instead of ``Sarcastic,'' where ``\<سار>'' is actually pronounced similarly to ``Sar''. In contrast, our fine-tuned versions of the model do not display such issues. 
This suggests that fine-tuning is critical for improving language-specific performance.


\section{Conclusion and Future Work}
\label{sec:conclusion}
In this study, we propose a specialized model, \llens{}, focused on news and social media analysis, designed to assist journalists, fact-checkers, and social media analysts. We curated 52 datasets covering Arabic, English, and Hindi, the key languages of the Arabian Peninsula. Using these, we built an instruction-following dataset and fine-tuned the Llama 3.1-8B-Instruct model for \llens{}.
Our experiments show that \llens{} outperforms the SOTA on 23 datasets, performs comparably on 8 datasets, and underperforms on the rest of the datasets. However, on average, \llens{} and its quantized versions significantly surpasses the Llama-instruct model. Our findings from error analysis suggests that it is important to inject specialized domain and language knowledge to obtain the desired outcome.
Our future studies include experimenting with different rank orders and focusing on quantized version of the model to make it usable in a low-resource settings.


\section{Limitations}
Our experiments were focused on a single open LLM, further LLMs can be explored. The training datasets had a bigger representation of Arabic, but as experiments showed, the proposed model improved performance even on other languages. Further examination of the effect of training examples selection and shuffling techniques is needed to understand these effects on the model performance. 

\section*{Ethics and Broader Impact}
Our experiments were conducted on training datasets publicly released to the research community. We adhered to the licenses associated with them whenever available. Some of the data points we will be releasing as part of our instruction dataset contain vulgar, offensive, or disturbing content which is a natural occurrence on social media, thus caution is recommended for users of our dataset. The models and instruction dataset we release could be invaluable to news agencies, journalists, and social media platforms. However, we encourage developers and users of the models to be critical of their usage.

\section*{Acknowledgments}
The work of M. Hasanain is supported by the NPRP grant 14C-0916-210015 from the Qatar National Research Fund part of Qatar Research Development and Innovation Council (QRDI). The findings achieved herein are solely the responsibility of the authors.

\bibliography{bib/bibliography}

\appendix
\section{Tasks and Datasets} 
\label{app:sec:datasets}

In this section, we provide a comprehensive overview of the tasks and datasets used throughout our study. Each task is outlined with a brief description, followed by the associated datasets that were utilized. The datasets are presented based on the specific languages and their relevance to the tasks at hand, with a focus on the binary and multi-labeled classifications where applicable. The tasks cover a wide range of objectives, from detecting harmful content to classifying news articles, identifying emotions, and more. This appendix serves as a detailed reference to support the methodology and scope of the research.

\subsection{Attentionworthiness}
\label{app:sec:attentionworthiness}

Attentionworthiness categorizes social media posts to determine whether they require attention and, if so, what kind of attention is needed. This task helps prioritize responses for policymakers by identifying critical posts that discuss actions, advice, or calls for intervention.

\noindent\paragraph{Dataset}
\label{app:sec:attentionworthiness_dataset}

For the Attentionworthiness Detection task, we utilized a subset of the Arabic dataset from Task 1D of the CLEF2022 CheckThat Lab~\cite{nakov2022overview1}. The SOTA number for this task was achieved by using Few-Shot GPT-4 (3-shot), as reported in LAraBench~\cite{abdelali-etal-2024-larabench}.

\subsection{Check-worthiness}
\label{app:sec:checkworthiness}

Check-worthiness helps streamline fact-checking by prioritizing claims most important for verification. This task operates as a binary classification, labeling tweets as either check-worthy or not check-worthy.

\noindent\paragraph{Dataset}
\label{app:sec:checkworthiness_dataset}

For the Check-worthiness task, we utilized both the English and Arabic subsets of the dataset released for Task 1 of the CLEF2024 CheckThat Lab~\cite{barron2024clef}. The dataset includes tweets labeled with binary classifications: check-worthy or not check-worthy. The SOTA for this task was achieved using GPT-3.5 (fine-tuned) for Arabic and RoBERTa for English, as reported in~\cite{struss2024overview}.

\subsection{Claim} 
Claim identifies whether a piece of text contains a factual, verifiable statement. A factual claim is one that can be substantiated through reliable sources, such as statistics, reports, or witness accounts. This task is crucial for fact-checking systems, as it helps distinguish between statements that can be objectively verified and those that are subjective or opinion-based.

\noindent\paragraph{Dataset} 
For the Claim task, we used three binary-labeled datasets: two in Arabic and one in English. The English dataset, ``claim-detection'' is sourced from Nithiwat/claim-detection on Hugging Face~\cite{nithiwat_claim_detection}. One of the Arabic datasets, ``CT22Claim'' comes from the Arabic subset of Task 1B of the CLEF2022 CheckThat Lab~\cite{nakov2022overview1}. The second Arabic dataset, ``ans-claim'' consists of true and false claims generated using crowdsourcing, based on the Arabic News Texts (ANT) corpus~\cite{khouja2005stance}. The SOTA for the Arabic dataset ``CT22Claim'' was achieved using Zero-shot GPT-3.5 in Larabench~\cite{abdelali-etal-2024-larabench}, while no SOTA was available for the English dataset ``claim-detection.''

\subsection{Cyberbullying} 
Cyberbullying identifies whether a piece of text contains abusive, harassing, or threatening behavior directed towards individuals online. This task plays a crucial role in moderating online spaces by flagging harmful content that can affect the well-being of users.

\noindent\paragraph{Dataset}
\label{app:sec:cyberbullying_dataset}

For the Cyberbullying task, we used datasets in three languages: Arabic, English, and Hindi. The Arabic dataset, ``ArCyc\_CB'' is sourced from~\citet{shannag2023arcyc}. The English dataset, ``Cyberbullying'' is developed by ~\citet{9378065}. The Hindi dataset, ``MC-Hinglish1.0'' is developed by ~\citet{rahman2024cyberbullying}. The SOTA for ArCyc\_CB was achieved using Support Vector Machine (SVM) with word embedding ~\cite{shannag2022offensive}. The English dataset ``Cyberbullying'' achieved its SOTA as reported in\cite{make6010009}, and for the Hindi dataset ``MC-Hinglish1.0,'' the SOTA was achieved using a voting classifier~\cite{rahman2024cyberbullying}.

\subsection{Emotion} 
Emotion  focuses on determining whether a piece of text conveys an emotion and identifying which specific emotion is being expressed.

\noindent\paragraph{Dataset}
\label{app:sec:emotion_detection_dataset}

For the Emotion task, we utilized two Arabic datasets and one English dataset. The Arabic datasets are ``Emotional-Tone''~\cite{al2018emotional} and ``NewsHeadline''~\cite{galal2022two}, while the English dataset is ``emotion''~\cite{saravia-etal-2018-carer}. The SOTA for the Emotional-Tone dataset was achieved using the Naïve Bayes Algorithm with 10-fold cross-validation~\cite{al2018emotional}. The NewsHeadline dataset achieved 100\% accuracy using Bag-of-Words (BOW) features~\cite{galal2022two}. For the English dataset, the CARER model was used to achieve the SOTA~\cite{saravia-etal-2018-carer}.

\subsection{Factuality} 
Factuality mainly focuses on assessing the truthfulness of a claim, determining whether the information presented is accurate or false.

\noindent\paragraph{Dataset} 
For the Factuality Detection task, we utilized five datasets: two in Arabic, two in English, and one in Hindi. The Arabic datasets are ``Arafacts''~\cite{ali2021arafacts} and ``COVID19Factuality''~\cite{alam-etal-2021-fighting-covid,alam2020call2arms}. The English datasets are ``News\_dataset''~\cite{dogo2020exploring} which combines two datasets (true and fake news) into one. The second English Dataset is ``Politifact''~\cite{Politifact}. Finally, the Hindi dataset is ``fake-news''~\cite{Patil-fakenews-hindi}. The datasets focus on classifying claims as either true or false to assess their factuality. The SOTA for ``COVID19Factuality'' by~\cite{alam-etal-2021-fighting-covid, alam2020call2arms}. The SOTA for ``News\_dataset'' was achieved using LSVM as a classifier~\cite{ahmed2017detection}, and for ``Politifact'' by~\cite{rangapur2023fin}. Moreover, Arafacts SOTA was obtained by ~\cite{othman2024arabic}. No SOTA was found for the fake-news in the Hindi dataset.

\subsection{Harmful} 
Harmful Detection focuses on determining whether a piece of text contains harmful content, which may include rumors, offensive language, hate speech, cyberbullying, violence, as well as racist, misogynistic, or sexist remarks. This task is essential for curbing the spread of harmful information online. The approach is based on work proposed in~\cite{alam-etal-2021-fighting-covid, nakov2022overview1}.

\noindent\paragraph{Dataset}
For the Harmful Detection task, we utilized the Arabic dataset ``CT22Harmful''~\cite{nakov2022overview1}, which is a binary dataset provided in Subtask 1C: Harmful tweet detection. The SOTA for this task was achieved by\cite{taboubi2022icompass}.

\subsection{Hate Speech} 
Hate Speech focuses on determining whether a piece of text contains hate speech. Hate speech refers to language that expresses hostility or animosity towards a specific group, or is intended to degrade, humiliate, or insult its members. This task is vital for monitoring online content and reducing the spread of harmful language.

\noindent\paragraph{Dataset} 
For the Hate Speech Detection task, we utilized four datasets: two in Arabic and two in Hindi. The Arabic datasets are ``annotated-hatetweets-4-classes''~\cite{ahmad2024arabic}, using the latest version 3, and ``OSACT4SubtaskB''~\cite{mubarak-etal-2020-overview}. The Hindi datasets are ``hate-speech-detection''~\cite{das2022hatecheckhin}, renamed from its original name ``HateCheckHIn'', and ``Hindi-Hostility-Detection-CONSTRAINT-2021''~\cite{bhardwaj2020hostility}, which is a multi-labeled dataset. The SOTA for ``annotated-hatetweets-4-classes'' was achieved by~\cite{ahmad2024hate}, for ``OSACT4SubtaskB'' by~\cite{husain2020osact4}, for ``hate-speech-detection'' by~\cite{das2022hatecheckhin}, and for ``Hindi-Hostility-Detection-CONSTRAINT-2021'' by~\cite{bhardwaj2020hostility}.

\subsection{Natural Language Inference}
Natural Language Inference focuses on identifying the relationship between two sentences. It involves determining whether the second sentence is logically supported by, contradicts, or remains neutral with respect to the first sentence.

\noindent\paragraph{Dataset} 
For the Natural Language Inference task, we used a binary-labeled Hindi dataset~\cite{gautam-etal-2021-translate}, which combines two datasets released by \citet{dhar-etal-2018-enabling} and \citet{srivastava-singh-2020-phinc}.  The SOTA for this task was achieved by~\cite{gautam2021translate}.

\subsection{News Credibility}
News Credibility involves determining whether a news article is reliable. This includes evaluating the article based on factors such as accuracy, fairness, objectivity, trustworthiness, completeness, and the presence or absence of biases.

\noindent\paragraph{Dataset}
For the News Credibility task, we used one Arabic dataset: ``NewsCredibilityDataset''~\cite{samdani2023arabic}.

\subsection{News Genre Categorization}
News Genre Categorization involves classifying news articles based on both their content and style. This task identifies the main topic or theme of an article, while also determining its genre, categorizing it as an opinion piece, objective news reporting, or satire.

\noindent\paragraph{Dataset} This task utilizes both Arabic and English datasets. The Arabic datasets include ASND''~\cite{chowdhury-etal-2020-improving-arabic}, ``SANADAkhbarona'', ``SANADAlArabiya'', ``SANADAlkhaleej''~\cite{einea2019sanad}, and ``UltimateDataset''~\cite{aldulaimi2022ultimate}. The English datasets include ``CNN\_News\_Articles\_2011-2022''~\cite{hadasu_cnn_web_crawler}, ``News\_Category\_Dataset''~\cite{misra2022news, misra2021sculpting}, with the latter containing 42 labels, and ``SemEval23T3-subtask1''~\cite{piskorski2023semeval}. The SOTA for the SANAD datasets was achieved by~\cite{elnagar2020arabic}, for ASND by~\cite{chowdhury2020improving}, for UltimateDataset by~\cite{setu2024empowering}, for CNN\_News\_Articles\_2011-2022 on Hugging Face by CHERGUELAINE Ayoub \& BOUBEKRI Faycal, and for SemEval23T3-subtask1 by~\cite{piskorski2023semeval}.

\subsection{Summarization}
Summarization focuses on producing concise and coherent summaries of articles, capturing the key points in a clear and succinct manner.

\noindent\paragraph{Dataset}
For the Summarization task, we used the "xlsum" multilingual dataset and the SOTA result ~\cite{hasan-etal-2021-xl} across all three languages: English, Arabic, and Hindi.

\subsection{Offensive Language}
Offensive Language Detection identifies whether a piece of text contains offensive language. Offensive speech includes vulgar or targeted insults, explicit or implicit attacks against others, or the use of inappropriate language.

\noindent\paragraph{Dataset}
Dataset for the Offensive Language Detection task, we used \textbf{five datasets}: two in Arabic, two in English, and one in Hindi. The Arabic datasets are ``ArCyc \_OFF'' (Shannag, 2023) and “OSACT4SubtaskA” (Zampieri et al., 2020). The English datasets are ``Offensive\_Hateful\_Dataset\_New'' (Christina, 2024) from Hugging Face, and ``offensive\_language\_dataset'' (Zampieri et al., 2019). The Hindi dataset is ``Offensive Speech Detection'' (Mathur et al., 2018). The labels for these datasets were extracted from Davidson et al. (2017). The SOTA for ``ArCyc\_OFF'' was achieved by~\cite{shannag2022offensive}, for ``Offensive Speech Detection'' by~\cite{das2022hatecheckhin}, for ``OSACT4SubtaskA'' by~\cite{mubarak2020overview}, and for ``offensive\_language\_dataset'' by~\cite{dehghani2024comprehensive}. The "hate-offensive-speech" dataset on Hugging Face achieved its SOTA by Purvesh Patel, while no SOTA was found for ``Offensive\_Hateful\_Dataset\_New''.



\subsection{Propaganda} 
Propaganda detection focuses on identifying propaganda in a piece of text. Propaganda is a form of communication designed to influence people’s opinions or actions toward a specific goal, often using strategic rhetorical and psychological techniques.

\noindent\paragraph{Dataset}
For the Propaganda task, we used two binary-labeled datasets: one in Arabic and one in English. The Arabic dataset and the SOTA which was achieved using AraBERT is ``ArPro''~\cite{hasanain2024can}, and the English dataset and SOTA is ``QProp''~\cite{barron2019proppy}. 

\subsection{Sarcasm}
Sarcasm focuses on determining whether a piece of text conveys sarcasm or not.

\noindent\paragraph{Dataset}
For the Sarcasm task, we used two binary-labeled datasets: one in Arabic and one in English. The Arabic dataset is ``ArSarcasm-v2''~\cite{abufarha-etal-2021-arsarcasm-v2}, and the English dataset is ``News-Headlines-Dataset-For-Sarcasm-Detection''~\cite{misra2023Sarcasm, misra2021sculpting}. The SOTA for both datasets is the same as the respective references where the datasets were found.

\subsection{Sentiment}
Sentiment classification involves in identifying and classifying sentiment expressed through text.

\noindent\paragraph{Dataset} 
For the Sentiment  task, we used four datasets: two in Arabic, one in English, and one in Hindi. The Arabic dataset ``ar\_reviews\_100k''~\cite{elnagar2018hotel, elnagar2016brad} includes hotel and book reviews from the HARD and BRAD datasets, with additional airline reviews. The second Arabic dataset is ``ArSAS''~\cite{elmadany2018arabic}. The English dataset is ``NewsMTSC-dataset''~\cite{Hamborg2021b}, and the Hindi dataset is ``Sentiment Analysis''~\cite{prabhu2016subword}. The SOTA for ``NewsMTSC-dataset'' was achieved by~\cite{Hamborg2021b}, and for ``Sentiment Analysis'' by~\cite{joshi2016towards}. No SOTA was found for ``ar\_reviews\_100k''.

\subsection{Stance}
Stance  involves predicting the author’s position toward a particular subject based on a written text. The stance may be expressed explicitly or implied within the content.

\noindent\paragraph{Dataset} 
For the Stance  task, we used two Arabic datasets: ``Mawqif-Arabic-Stance-main''~\cite{alturayeif-etal-2022-mawqif} and ``stance''~\cite{khouja2005stance}. The SOTA for ``Mawqif-Arabic-Stance-main'' was achieved by~\cite{alturayeif2022mawqif}, and for ``stance'', the SOTA was achieved using pretraining (BERT), as reported in the same dataset reference~\cite{khouja2005stance}.

\subsection{Subjectivity} 
Subjectivity involves determining whether a piece of text is subjective or objective. A sentence is considered subjective if it is influenced by personal feelings, tastes, or opinions; otherwise, it is classified as objective.

\noindent\paragraph{Dataset}
For the Subjectivity task, we used two binary-labeled datasets: one in Arabic and one in English. The Arabic dataset is ``ThatiAR''~\cite{suwaileh2024thatiarsubjectivitydetectionarabic}, and the English dataset is ``CT24\_T2'' from the CLEF2024--CheckThat-Lab Task 2~\cite{clef-checkthat:2024-lncs}. The SOTA for ``ThatiAR'' was achieved by~\cite{suwaileh2024thatiarsubjectivitydetectionarabic}, and the SOTA for ``CT24\_T2'' was achieved by~\cite{gruman2024clac}.

\section{Dataset Sizes After Pre-processing}
\label{app:sec:datasets_sizes}
The Tables \ref{tab:distribution_arabic_dataset}, \ref{tab:distribution_english_dataset} and \ref{tab:distribution_hindi_dataset} present the sizes of the datasets used in this study, after pre-processing. The datasets are categorized by language (Arabic, English, and Hindi) and we report the distribution of training, development test sets, along with the number of labels for each task.

\begin{table*}[h]
\centering
\resizebox{0.9\textwidth}{!}{%
\begin{tabular}{lllrlr}
\toprule
\textbf{Task} & \textbf{Dataset} & \textbf{\# Labels} & \textbf{\# Train}  &\textbf{\# Dev} 
& \textbf{\# Test} \\
\midrule
Attentionworthiness & CT22Attentionworthy & 9 
& 2,470  &1,071 
& 1,186 \\
Checkworthiness &CT24\_T1 & 2 
& 22,403  &1093
& 500\\
Claim & CT22Claim & 2 
& 3,513  &339 
& 1,248 \\
Cyberbullying & ArCyc\_CB & 2 
& 3,145  &451 
& 900 \\
Emotion  & Emotional-Tone & 8 
& 7,024  &1,005 
& 2,009 \\
Emotion  & NewsHeadline & 7 
& 939  &160 
& 323 \\
Factuality & Arafacts & 5 
& 4,354  &623 
& 1,245 \\
Factuality & COVID19Factuality & 2 
& 3,513  &339 
& 988 \\
Harmful & CT22Harmful & 2 
& 2,484  &1,076 
& 1,201 \\
Hate Speech & annotated-hatetweets-4-classes & 4 
& 210,526  &90,544 
& 100,565 \\
Hate Speech & OSACT4SubtaskB & 2 
& 4,778  &2,048 
& 1,827 \\
News Genre Categorization & ASND & 10 
& 74,496  &11,136 
& 21,942 \\
News Genre Categorization & SANADAkhbarona & 7 
& 62,210  &7,824 
& 7,824 \\
News Genre Categorization & SANADAlArabiya & 6 
& 56,967  &7,120 
& 7,123 \\
News Genre Categorization & SANADAlkhaleej & 7 
& 36,391  &4,550 
& 4,550 \\
News Genre Categorization & UltimateDataset & 10 & 133,036  &19,269 
& 38,456 \\
News Credibility & NewsCredibilityDataset & 2 & 8,671  &1,426 
& 2,730 \\
Summarization & xlsum & -- & 37,425  &4,689 
& 4,689 \\
Offensive Language & ArCyc\_OFF & 2 & 3,138  &450 
& 900 \\
Offensive Language & OSACT4SubtaskA & 2 & 4,780  &2,047 
& 1,827 \\
Propaganda & ArPro & 2 & 6,002  &672 
& 1,326 \\
Sarcasm & ArSarcasm-v2 & 2 & 8,749  &3,761 
& 2,996 \\
Sentiment  & ar\_reviews\_100k & 3 & 69,998  &10,000 
& 20,000 \\
Sentiment  & ArSAS & 4 & 13,883  &1,987 
& 3,976 \\
Stance  & Mawqif-Arabic-Stance-main & 2 & 3,162  &950 
& 560 \\
Stance  & stance & 3 & 2,652  &755 
& 379 \\
Subjectivity & ThatiAR & 2 & 2,446  &467 & 748 \\
\bottomrule
\end{tabular}%
}
\caption{Data distribution across Arabic datasets.}
\label{tab:distribution_arabic_dataset}
\end{table*}


\begin{table*}[h]
\centering
\resizebox{\textwidth}{!}{%
\begin{tabular}{lllrlr}
\toprule
\textbf{Task} & \textbf{Dataset} & \textbf{\# Labels} & \textbf{\# Train}  &\textbf{\# Dev} 
& \textbf{\# Test} \\
\midrule
Checkworthiness &CT24\_T1 & 2 & 22,403  &318 
& 1,031 \\
Claim & claim-detection & 2 & 23,224  &5,815 
& 7,267 \\
Cyberbullying & Cyberbullying & 6 & 32,551  &4,751 
& 9,473 \\
Emotion  & emotion & 6 & 280,551  &41,429 
& 82,454 \\
Factuality & News\_dataset & 2 & 28,147  &4,376 
& 8,616 \\
Factuality & Politifact & 6 & 14,799  &2,116 
& 4,230 \\
News Genre Categorization & CNN\_News\_Articles\_2011-2022 & 6 & 32,193  &9,663 
& 5,682 \\
News Genre Categorization & News\_Category\_Dataset & 42 & 145,748  &20,899 
& 41,740 \\
News Genre Categorization & SemEval23T3-subtask1 & 3 & 302  &130 
& 83 \\
Summarization & xlsum & -- & 306,493  &11,535 
& 11,535 \\
Offensive Language & Offensive\_Hateful\_Dataset\_New & 2 & 42,000  &5,254 
& 5,252 \\
Offensive Language & offensive\_language\_dataset & 2 & 29,216  &3,653 
& 3,653 \\
Offensive/Hate-Speech & hate-offensive-speech & 3 & 48,944  &2,802 
& 2,799 \\
Propaganda & QProp & 2 & 35,986  &5,125 
& 10,159 \\
Sarcasm & News-Headlines-Dataset-For-Sarcasm-Detection & 2 & 19,965  &2,858 
& 5,719 \\
Sentiment  & NewsMTSC-dataset & 3 & 7,739  &320 
& 747 \\
Subjectivity & clef2024-checkthat-lab & 2 & 825  &219 & 484 \\
\bottomrule
\end{tabular}%
}
\caption{Data distribution across English datasets.}
\label{tab:distribution_english_dataset}
\end{table*}

\begin{table*}[h]
\centering
\resizebox{\textwidth}{!}{%
\begin{tabular}{lllrlr}
\toprule
\textbf{Task} & \textbf{Dataset} & \textbf{\# Labels} & \textbf{\# Train}  &\textbf{\# Dev} 
& \textbf{\# Test} \\
\midrule
Cyberbullying & MC-Hinglish1.0 & 7 & 7,400  &318 
& 1,000 \\
Factuality & fake-news & 2 & 8,393  &5,815 
& 2,743 \\
Hate Speech & hate-speech-detection & 2 & 3,327  &4,751 
& 951 \\
Hate Speech & Hindi-Hostility-Detection-CONSTRAINT-2021 & 15 & 5,718  &41,429 
& 1,651 \\
Natural Language Inference & Natural Language Inference & 2 & 1,251  &4,376 
& 447 \\
Summarization & xlsum & -- & 70,754  &2,116 
& 8,847 \\
Offensive Speech & Offensive Speech Detection & 3 & 2,172  &9,663 
& 636 \\
Sentiment  & Sentiment Analysis & 3 & 10,039  &20,899 
& 1,259 \\
\bottomrule
\end{tabular}%
}
\caption{Data distribution across Hindi datasets.}  
\label{tab:distribution_hindi_dataset}
\end{table*}

\section{Instructions Generation}
\label{app:sec:app:prompting_inst}
For each task, each dataset and each language, we use two effective closed models, GPT-4o and Claude-3.5 Sonnet, to generate instructions. These instructions were used to create the final instruction dataset for LLM fine-tuning. We prompt the models to generate these instructions using the prompt in Table~\ref{tab:prompts-inst-gen}. For all generated instructions, we append the 
following suffix to further instruct the LLM to limit its responses to the labels/summary, to simplify post-processing at inference time: \textit{Return only the label without any explanation, justification or additional text}. Table~\ref{tab:ex_insts} shows examples of the generated instructions. Note that we only generate instructions for the user role, while we keep the system role fixed to that presented in Table~\ref{tab:ex_insts}. 

\begin{table*}[h]
\centering
\resizebox{\textwidth}{!}{%
\begin{tabular}{lp{14cm}}
\toprule
\textbf{Role} & \textbf{Prompt}  \\
\midrule
System & You are an expert LLM developer with expertise in writing instructions to instruction-tune LLMs for users` tasks.\\
\midrule
User  & We are creating an \textit{[INSTRUCT-LANG]} instruction-following dataset for a/an [\textit{LANG}] dataset called: [\textit{DATASET}] covering the task of [\textit{TASK}]. The user defined the task as follows: [\textit{TASK DEFINITION}]. For that task, the labels include: [\textit{LABELS}]. Write 10 very diverse and concise English instructions. Return the instructions as strings in a list format as follows [].\\
\bottomrule
\end{tabular}%
}
\caption{Prompts used to generate instructions through LLMs. \textit{INSTRUCT-LANG} refers to the language, which can be Arabic, English, or Hindi. \textit{LANG} also denotes the language, specifically Arabic, English, or Hindi. \textit{TASK} refers to the task name. For each task, there is a \textit{TASK DEFINITION}. \textit{LABELS} refers to dataset-specific labels.
\label{tab:prompts-inst-gen}
}
\end{table*}

\begin{table*}[h]
\centering
\resizebox{\textwidth}{!}{%
\begin{tabular}{lp{8cm}p{5.5cm}}
\toprule
\textbf{Model} & \textbf{Instruction}  & \textbf{System Role}  \\
\midrule
GPT-4o & Classify the given text as either `offensive' or `not-offensive-hateful'. Return only the label without any explanation, justification or additional text.&  You are a social media expert providing accurate analysis and insights. \\
\midrule
Claude-3.5  & Evaluate whether the given text is `offensive' or `not-offensive-hateful', considering vulgar or targeted attacks. Return only the label without any explanation, justification or additional text. & You are a social media expert providing accurate analysis and insights. \\
\bottomrule
\end{tabular}%
}
\caption{Examples of instructions generated by two LLMs for the offensive language detection task on the English offensive\_language\_dataset, along with the pre-defined system role prompt.\label{tab:ex_insts}}
\end{table*}

\section{Data Release and License}
\label{sec:app_data_release}
The \llens{} model and the instruction-following dataset will be publicly released under the Creative Commons Attribution Non Commercial Share Alike 4.0: \url{https://creativecommons.org/licenses/by-nc-sa/4.0/}.

\section{Use of AI assistant}
\label{sec:use_of_ai_assistant}
We used AI assistants such as GPT-4o and Claude for generating the instructions dataset, as well as for spelling and grammar checking for the text of the paper.

\begin{table*}[!ht]
\centering
\footnotesize  
\renewcommand{\arraystretch}{1.1}  
\setlength{\tabcolsep}{3pt}  

\begin{tabular}{@{}lllrccc@{}}
    \toprule
    \textbf{Task} & \textbf{Dataset} & \textbf{Metric} 
    & \textbf{SOTA} & \textbf{Base} & \textbf{L-Lens (Eng.)} & \textbf{L-Lens (Native)} \\
    \midrule
    \multicolumn{7}{c}{\textbf{Natural Language Understanding (NLU)}} \\
    \midrule
    \textbf{Overall} & ~ & ~ & 0.137 & 0.034 & 0.129 & 0.130 \\
        \midrule

    News Sum & xlsum & R-2 & 0.137 & 0.034 & 0.129 & 0.130 \\
    \midrule

    \multicolumn{7}{c}{\textbf{Information Extraction \& Text Classification}} \\
    \midrule
    \textbf{Overall} & ~ & ~ & 0.923 & 0.660 & 0.869 & 0.864 \\
        \midrule

    News Cat & ASND & Ma-F1 & 0.770 & 0.587 & 0.919 & 0.929 \\
    News Cat & SANADAkhbarona & Acc & 0.940 & 0.784 & 0.954 & 0.953 \\
    News Cat & SANADAlArabiya & Acc & 0.974 & 0.893 & 0.987 & 0.985 \\
    News Cat & SANADAlkhaleej & Acc & 0.986 & 0.865 & 0.984 & 0.982 \\
    News Cat & UltimateDataset & Ma-F1 & 0.970 & 0.376 & 0.865 & 0.880 \\
    News Credibility & NewsCredibility & Acc & 0.899 & 0.455 & 0.935 & 0.933 \\
    Subjectivity & ThatiAR & F1\_Pos & 0.800 & 0.562 & 0.441 & 0.383 \\
    \midrule

    \multicolumn{7}{c}{\textbf{Emotion \& Sentiment Analysis}} \\
    \midrule
    \textbf{Overall} & ~ & ~ & 0.786 & 0.557 & 0.724 & 0.718 \\
        \midrule

    Emotion & Emotional-Tone & W-F1 & 0.658 & 0.358 & 0.705 & 0.736 \\
    Emotion & NewsHeadline & Acc & 1.000 & 0.406 & 0.480 & 0.458 \\
    Sarcasm & ArSarcasm-v2 & F1\_Pos & 0.584 & 0.477 & 0.520 & 0.542 \\
    Sentiment & ar\_reviews\_100k & F1\_Pos & -- & 0.681 & 0.785 & 0.779 \\
    Sentiment & ArSAS & Acc & 0.920 & 0.603 & 0.800 & 0.804 \\
    Stance & stance & Ma-F1 & 0.767 & 0.608 & 0.926 & 0.881 \\
    Stance & Mawqif-Arabic-Stance & Ma-F1 & 0.789 & 0.764 & 0.853 & 0.826 \\
    \midrule

    \multicolumn{7}{c}{\textbf{Fast-Checking \& Verification}} \\
    \midrule
    \textbf{Overall} & ~ & ~ & 0.689 & 0.407 & 0.663 & 0.671 \\
        \midrule
    Att.worthiness & CT22Attentionworthy & W-F1 & 0.412 & 0.158 & 0.425 & 0.454 \\
    Checkworthiness & CT24\_T1 & F1\_Pos & 0.569 & 0.610 & 0.502 & 0.509 \\
    Claim & CT22Claim & Acc & 0.703 & 0.581 & 0.734 & 0.756 \\
    Factuality & Arafacts & Mi-F1 & 0.850 & 0.210 & 0.771 & 0.738 \\
    Factuality & COVID19Factuality & W-F1 & 0.831 & 0.492 & 0.800 & 0.840 \\
    Propaganda & ArPro & Mi-F1 & 0.767 & 0.597 & 0.747 & 0.731 \\
    \midrule

    \multicolumn{7}{c}{\textbf{Hate Speech \& Offensive Content}} \\
    \midrule
    \textbf{Overall} & ~ & ~ & 0.797 & 0.603 & 0.774 & 0.767 \\
        \midrule

    Cyberbullying & ArCyc\_CB & Acc & 0.863 & 0.766 & 0.870 & 0.833 \\
    Harmfulness & CT22Harmful & F1\_Pos & 0.557 & 0.507 & 0.523 & 0.535 \\
    Hate Speech & annotated-hatetweets-4 & W-F1 & 0.630 & 0.257 & 0.526 & 0.517 \\
    Hate Speech & OSACT4SubtaskB & Mi-F1 & 0.950 & 0.819 & 0.955 & 0.955 \\
    Offensive & ArCyc\_OFF & Ma-F1 & 0.878 & 0.489 & 0.877 & 0.879 \\
    Offensive & OSACT4SubtaskA & Ma-F1 & 0.905 & 0.782 & 0.896 & 0.882 \\
    \bottomrule
\end{tabular}

\caption{Arabic Capability-Based Evaluation Results. Base: Llama3.1-Instruct. L-Lens: LlamaLens. Eng.: English. Acc: Accuracy. W-F1: Weighted F1. Mi-F1: Micro-averaged F1. Ma-F1: Macro-averaged F1, R-2: ROUGE-2. Cat:Categorization. News Sum: News Summarization. Att.worthiness: Attentionworthiness.  }
\label{Arabic-Capability-Based}
\end{table*}

\clearpage
\begin{table*}[!ht]
\centering
\footnotesize  
\renewcommand{\arraystretch}{1.1}  
\setlength{\tabcolsep}{3pt}  

\begin{tabular}{@{}lllrccc@{}}
    \toprule
    \textbf{Task} & \textbf{Dataset} & \textbf{Metric} 
    & \textbf{SOTA} & \textbf{Base} & \textbf{\shortstack{L-Lens \\ (Eng.)}} & \textbf{\shortstack{L-Lens \\ (Native)}} \\
    \midrule
    \multicolumn{7}{c}{\textbf{English Capability-Based Evaluation}} \\
    \midrule
    \textbf{Overall} & ~ & ~ & 0.152 & 0.074 & 0.182 & 0.181
\\
    \midrule
    News Sum & xlsum & R-2 & 0.152 & 0.074 & 0.182 & 0.181 \\
    \midrule
    \multicolumn{7}{c}{\textbf{Information Extraction \& Text Classification}} \\
    \midrule
    \textbf{Overall} & ~ & ~ & 0.817 & 0.709 & 0.669 & 0.593 \\
    \midrule
    News Cat & CNN\_News\_Articles & Acc & 0.940 & 0.644 & 0.970 & 0.970 \\
    News Cat & News\_Category & Ma-F1 & 0.769 & 0.970 & 0.824 & 0.520 \\
    News Genre & SemEval23T3-ST1 & Mi-F1 & 0.815 & 0.687 & 0.241 & 0.253 \\
    Subjectivity & CT24\_T2 & Ma-F1 & 0.744 & 0.535 & 0.642 & 0.628 \\
    \midrule
    \multicolumn{7}{c}{\textbf{Emotion \& Sentiment Analysis}} \\
    \midrule
    \textbf{Overall} & ~ & ~ & 0.835 & 0.550 & 0.830 & 0.834 \\
    \midrule
    Emotion & emotion & Ma-F1 & 0.790 & 0.353 & 0.803 & 0.808 \\
    Sarcasm & News-Headlines & Acc & 0.897 & 0.668 & 0.936 & 0.947 \\
    Sentiment & NewsMTSC & Ma-F1 & 0.817 & 0.628 & 0.751 & 0.748 \\
    \midrule
    \multicolumn{7}{c}{\textbf{Fact-Checking \& Verification}} \\
    \midrule
    \textbf{Overall} & ~ & ~ & 0.708 & 0.497 & 0.811 & 0.823 \\
    \midrule
    Checkworthiness & CT24\_T1 & F1\_Pos & 0.753 & 0.404 & 0.942 & 0.942 \\
    Claim & claim-detection & Mi-F1 & -- & 0.545 & 0.864 & 0.889 \\
    Factuality & News\_dataset & Acc & 0.920 & 0.654 & 0.999 & 0.999 \\
    Factuality & Politifact & W-F1 & 0.490 & 0.121 & 0.287 & 0.311 \\
    Propaganda & QProp & Ma-F1 & 0.667 & 0.759 & 0.963 & 0.973 \\
    \midrule
    \multicolumn{7}{c}{\textbf{Hate Speech \& Offensive Content}} \\
    \midrule
    \textbf{Overall} & ~ & ~ & 0.949 & 0.529 & 0.870 & 0.874 \\
    \midrule
    Cyberbullying & Cyberbullying & Acc & 0.907 & 0.175 & 0.836 & 0.855 \\
    Offensive & Offensive\_Hateful & Mi-F1 & -- & 0.692 & 0.814 & 0.813 \\
    Offensive & offensive\_language & Mi-F1 & 0.994 & 0.646 & 0.899 & 0.893 \\
    Offensive \& Hate & hate-offensive-speech & Acc & 0.945 & 0.602 & 0.931 & 0.935 \\
    \midrule
    \multicolumn{7}{c}{\textbf{Hindi Capability-Based Evaluation}} \\
    \midrule
    \textbf{Overall} & ~ & ~ & 0.391 & 0.356 & 0.369 & 0.425 \\
    \midrule
    NLI & NLI\_dataset & W-F1 & 0.646 & 0.633 & 0.568 & 0.679 \\
    News Sum & xlsum & R-2 & 0.136 & 0.078 & 0.171 & 0.170 \\
    \midrule
    \multicolumn{7}{c}{\textbf{Emotion \& Sentiment Analysis}} \\
    \midrule
    \textbf{Overall} & ~ & ~ & 0.697 & 0.552 & 0.647 & 0.654 \\
    \midrule
    Sentiment & Sentiment Analysis & Acc & 0.697 & 0.552 & 0.647 & 0.654 \\
    \midrule
    \multicolumn{7}{c}{\textbf{Fact-Checking \& Verification}} \\
    \midrule
    \textbf{Overall} & ~ & ~ & -- & 0.759 & 0.994 & 0.993 \\
    \midrule
    Factuality & fake-news & Mi-F1 & -- & 0.759 & 0.994 & 0.993 \\
    \midrule
    \multicolumn{7}{c}{\textbf{Hate Speech \& Offensive Content}} \\
    \midrule
    \textbf{Overall} & ~ & ~ & 0.703 & 0.518 & 0.801 & 0.802 \\
    \midrule
    Hate Speech & hate-speech-detection & Mi-F1 & 0.639 & 0.750 & 0.963 & 0.963 \\
    Hate Speech & Hindi-Hostility & W-F1 & 0.841 & 0.469 & 0.753 & 0.753 \\
    Offensive & Offensive Speech & Mi-F1 & 0.723 & 0.621 & 0.862 & 0.865 \\
    Cyberbullying & MC\_Hinglish1 & Acc & 0.609 & 0.233 & 0.625 & 0.627 \\
    \bottomrule
\end{tabular}

\caption{Capability-Based Evaluation Results for English and Hindi. Base: Llama3.1-Instruct. L-Lens: LlamaLens. Eng.: English. Acc: Accuracy. W-F1: Weighted F1. Mi-F1: Micro-averaged F1. Ma-F1: Macro-averaged F1, R-2: ROUGE-2. News Sum: News Summarization. Cat:Categorization. NLI: Natural Language Inference.}
\label{English-Hindi-Capability-Based}
\end{table*}

\clearpage

\end{document}